\documentclass{article}
\usepackage{ijcai23}

\usepackage{times}
\usepackage{soul}
\usepackage{url}
\usepackage[hidelinks]{hyperref}
\usepackage[utf8]{inputenc}
\usepackage[small]{caption}
\usepackage{graphicx}
\usepackage{amsmath}
\usepackage{amsthm}
\usepackage{booktabs}
\usepackage{algorithm}
\usepackage{algorithmic}
\usepackage[switch]{lineno}

\title{TherapyView: Visualizing Therapy Sessions with Temporal Topic Modeling and AI-Generated Arts}

\author{
Baihan Lin$^1$, Stefan Zecevic$^2$, Djallel Bouneffouf$^2$, Guillermo Cecchi$^2$\\
\affiliations
$^1$ Columbia University, New York, NY 10027, USA\\
$^2$ IBM TJ Watson Research Center, Yorktown Heights, NY 10598, USA\\
\emails
baihan.lin@columbia.edu, \{szecevic, djallel.bouneffouf\}@ibm.com, gcecchi@us.ibm.com
}

\begin{document}

\maketitle

\begin{abstract}
We present the TherapyView, a demonstration system to help therapists visualize the dynamic contents of past treatment sessions, enabled by the state-of-the-art neural topic modeling techniques to analyze the topical tendencies of various psychiatric conditions and deep learning-based image generation engine to provide a visual summary. The system incorporates temporal modeling to provide a time-series representation of topic similarities at a turn-level resolution and AI-generated artworks given the dialogue segments to provide a concise representations of the contents covered in the session, offering interpretable insights for therapists to optimize their strategies and enhance the effectiveness of psychotherapy. This system provides a proof of concept of AI-augmented therapy tools with e in-depth understanding of the patient's mental state and enabling more effective treatment.
\end{abstract}

\section{Introduction}

Mental health is a global issue affecting people of all ages, cultures, and countries. 
With the increasing demand for mental health services, innovative solutions are needed to address the workload on mental health providers. One promising area of innovation is natural language processing (NLP), which has been adopted in psychotherapy to help therapists provide better care \cite{shum2018eliza,zemvcik2019brief,rezaii2022natural}. Previous works have demonstrated the effectiveness of classical topic modeling in mental illness detection \cite{resnik2015beyond,zeng2012synonym}, but recent advancements in deep learning have led to the emergence of neural topic modeling as a better solution \cite{lin2022neural}.

In this paper, we propose TherapyView, a data visualization system that uses neural topic modeling to learn the topical propensities of different psychiatric conditions from psychotherapy session transcripts. We conducted an empirical evaluation of existing neural topic modeling techniques with a focus on their application to the domain of psychotherapy, benchmarked on the Alexander Street Counseling and Psychotherapy Transcripts dataset. By leveraging temporal modeling of state-of-the-art topic models, we provide a visual representation of the topical tendencies of psychiatric conditions, enabling therapists to easily identify patterns and make informed decisions about their psychotherapy strategies.

Moreover, we introduce AI-generated arts on different temporal segments of the therapy sessions, providing a concise visual summary of the session. The user-friendly interface and interactive visualizations make it easy for therapists to understand and interpret the results, leading to improved treatment outcomes for patients. Our data visualization system is a powerful tool for advancing the field of psychotherapy and providing therapists with the information they need to make informed decisions.

\begin{figure}[tb]
\centering
    \includegraphics[width=\linewidth]{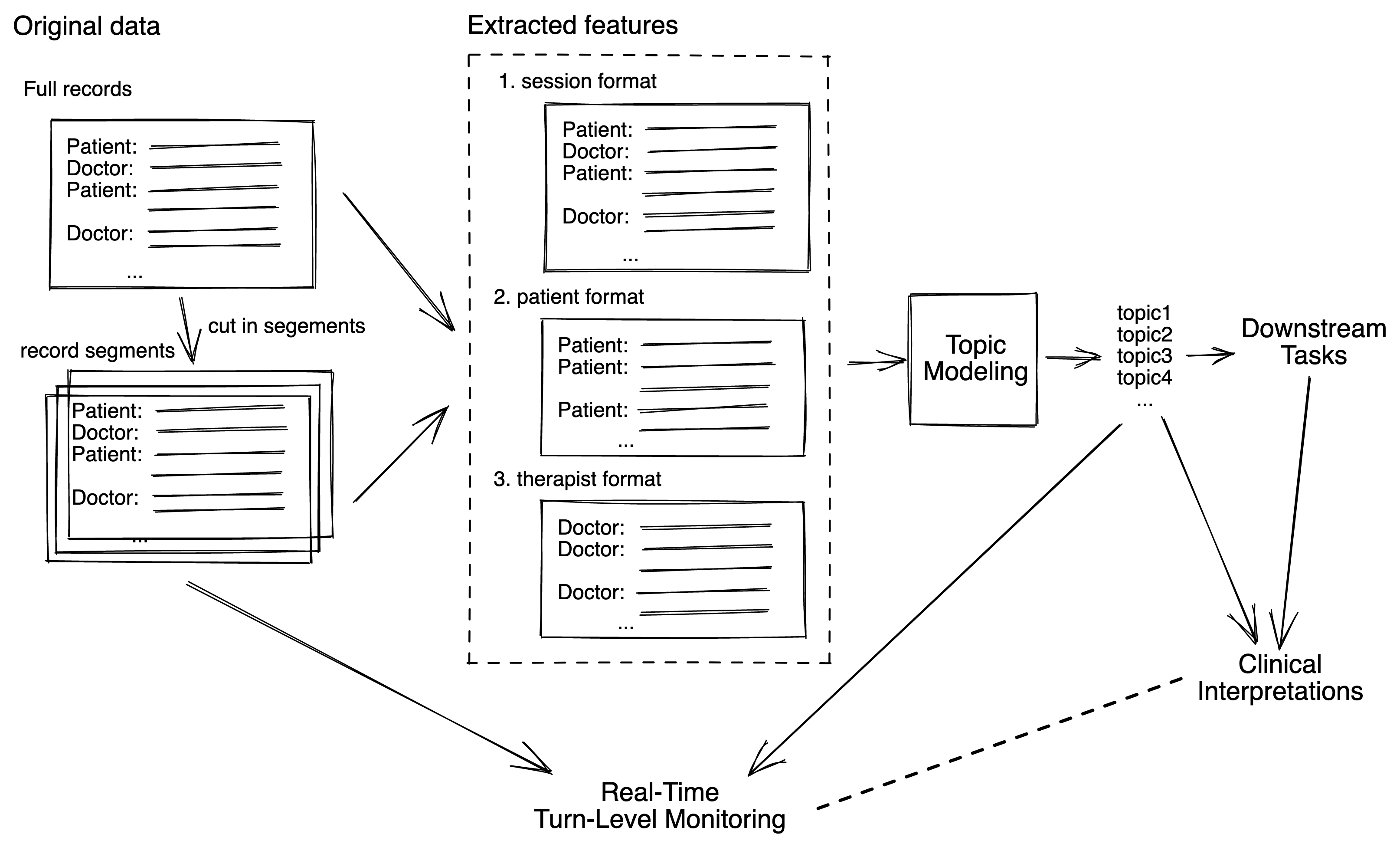}
\caption{Psychotherapy topic modeling framework
}\label{fig:pipeline}
\end{figure}

\section{The Temporal Topic Modeling Framework}



\begin{algorithm}[tb]
 \caption{Temporal Topic Modeling (TTM)}
 \label{alg:ttm}
 \begin{algorithmic}[1]
 \STATE {\bfseries } Learned topics $T$ as references
 \STATE {\bfseries }\textbf{for} i = 1,2,$\cdots$, N \textbf{do}
\STATE {\bfseries } \quad Automatically transcribe dialogue turn pairs  $(S^p_i,S^t_i)$
\STATE {\bfseries }\quad \textbf{for} $T_j \in$ topics $T$ \textbf{do} \\
\STATE {\bfseries }\quad \quad Topic score $W^{p_i}_{j}$ = similarity($Emb({T_j}), Emb(S^p_i)$) \\
\STATE {\bfseries }\quad \quad Topic score $W^{t_i}_{j}$ = similarity($Emb({T_j}), Emb(S^t_i)$) \\
\STATE {\bfseries } \quad \textbf{end for}
\STATE {\bfseries } \textbf{end for}
 \end{algorithmic}
\end{algorithm}

Our proposed analytical framework, outlined in Figure \ref{fig:pipeline}, leverages natural language processing (NLP) techniques and neural topic modeling to extract valuable insights from psychotherapy session transcripts. During the session, the dialogue between the patient and therapist is transcribed into pairs of turns, which are then used as the input data for our framework. We can take the full records of a patient or a cohort of patients belonging to the same condition, which can either be used as is before feature extraction or truncated into segments based on timestamps or topic turns.

We extract features from the transcript data using NLP techniques and fit them into neural topic models to generate a list of weighted topic words. These topic words provide important insights into the patient's condition and are often highly interpretable, making them valuable in the context of psychotherapy.

Our framework offers a range of downstream tasks and user scenarios. We can use the extracted weighted topics to assess the progress of the therapy, identify potential issues in the patient's mental state, or suggest adjustments to the therapist's treatment strategies. These features can be incorporated into an intelligent AI assistant to help remind the therapist of important information during the session. Additionally, certain taboo topics such as those related to suicidal conversations can be flagged for the therapist's attention.

To further analyze the transcript data, we use temporal topic modeling (TMM) to compute turn-resolution topic scores. Algorithm \ref{alg:ttm} outlines the pipeline of our TMM analysis. For example, if we have learned ten topics, the topic score will be a ten-dimensional vector, with each dimension corresponding to a likelihood of the turn being in that topic. To account for the directional property of each turn with respect to a given topic, we compute the cosine similarity between the embedded topic vector and the embedded turn vector, instead of directly inferring the probability as in traditional topic assignment problems. The Embedded Topic Model (ETM), which we use for temporal modeling in the results section, also models each word with a categorical distribution whose natural parameter is the inner product between a word embedding and an embedding of its assigned topic. We use Word2Vec \cite{mikolov2013distributed} as our word embedding for both the topics and the turns.

\section{The Empirical Evaluations}


\begin{table}[t]
      \caption{Topic evaluations of the neural topic models
      }
      \label{tab:eval} 
      \centering
      \resizebox{\linewidth}{!}{
 \begin{tabular}{ l | c | c | c | c | c | c |  }
 &\multicolumn{2}{c}{Anxiety} \vline &\multicolumn{2}{c}{Depression} \vline &\multicolumn{2}{c}{Schizophrenia} \vline \\
  & TC & TD  & TC & TD   & TC & TD   \\ \hline
NVDM-GSM \cite{miao2017discovering} & 0.653 & \textbf{-380.933} & 0.487 & -316.439 & 0.527 & -431.393 \\
WTM-MMD \cite{nan2019topic} & \textbf{0.927} & -453.929 & 0.907 & -359.964 & 0.447 & -403.694 \\
ETM \cite{dieng2020topic} & 0.893 & -449.000 & \textbf{0.933} & -367.069 &  \textbf{0.973} & -310.211 \\
BATM \cite{wang2020neural} & 0.720 & -441.049 & 0.773 & -443.394 & 0.500 & -337.825 \\
\end{tabular}
 }
\end{table}

To evaluate the performance of our proposed neural topic modeling approach, we compare four state-of-the-art models and analyze their learned topics on our dataset \cite{miao2017discovering,nan2019topic,dieng2020topic,wang2020neural}. We separate the transcript sessions into three categories based on the psychiatric conditions of the patients (anxiety, depression, and schizophrenia), and train the topic models for over 100 epochs at a batch size of 16. As with standard preprocessing for topic modeling, we set the lower bound of the word count to keep in topic training to be 3, and the ratio of the upper bound of the word count to keep in topic training to be 0.3. We evaluate the models using a series of validated measurements of topic coherence and diversity, as outlined in \cite{roder2015exploring}. Specifically, we use an asymmetrical confirmation measure between top word pairs (smoothed conditional probability) for topic coherence and the ratio between the size of the vocabulary in the topic words and the total number of words in the topics for topic diversity.

Table \ref{tab:eval} summarizes the evaluation results for the four models across the different psychiatric conditions, based on validated measurements of topic coherence and diversity proposed in \cite{roder2015exploring}. We observe that the Embedded Topic Model (ETM) yields relatively high topic coherence and diversity across all three psychiatric conditions in the Alex Street datasets, making it a suitable choice for our deployed system. For the full evaluation results, see \cite{lin2022neural}.

To ensure that the learned topics can be mapped from one clinical condition to another, we compute a universal topic model on the text corpus of the entire Alex Street psychotherapy database. Using this universal topic model, we compute a 10-dimensional topic score for each turn, corresponding to the 10 topics. The higher the score, the more positively correlated the turn is with the topic. This time-series matrix enables us to probe the dynamics of the dialogues within the topic space (e.g. visualized as a 3d trajectory in our demonstration system, TherapyView).

To provide interpretable insights, it is important to parse out the concepts behind the learned topics. To better understand the topics, we parse out the highest-scoring turns in the transcripts that correspond to each topic. For example, topic 0 is about figuring out self-discovery and reminiscence, while topic 1 is about play. Topic 2 is about anger, fear, and sadness, while topic 3 is about counts. Topic 4 is about tiredness and decision-making, while topic 5 is about sickness, self-injuries, and coping mechanisms. Topic 6 is about explicit ways to deal with stress, such as keeping busy and reaching out for help, while topic 7 is about numbers. Topic 8 is about continuation and perseverance, while topic 9 is mostly chit-chat, interjections, and transcribed prosody.

\begin{figure}[tb]
\centering
    \includegraphics[width=\linewidth]{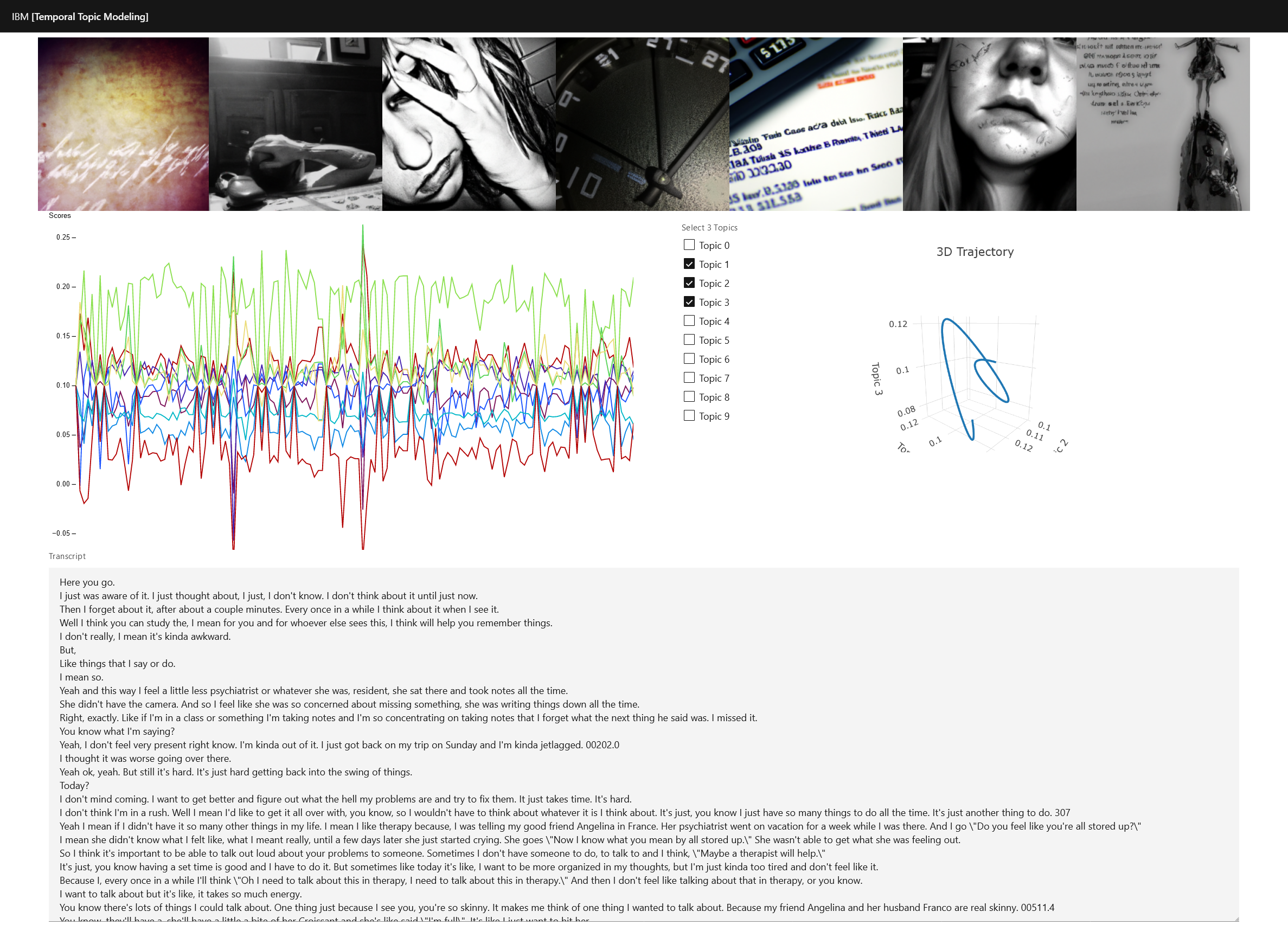}
\caption{Screenshot of the launch page of TherapyView dashboard
}\label{fig:screenshot}
\end{figure}

\begin{figure}[tb]
\centering
    \includegraphics[width=\linewidth]{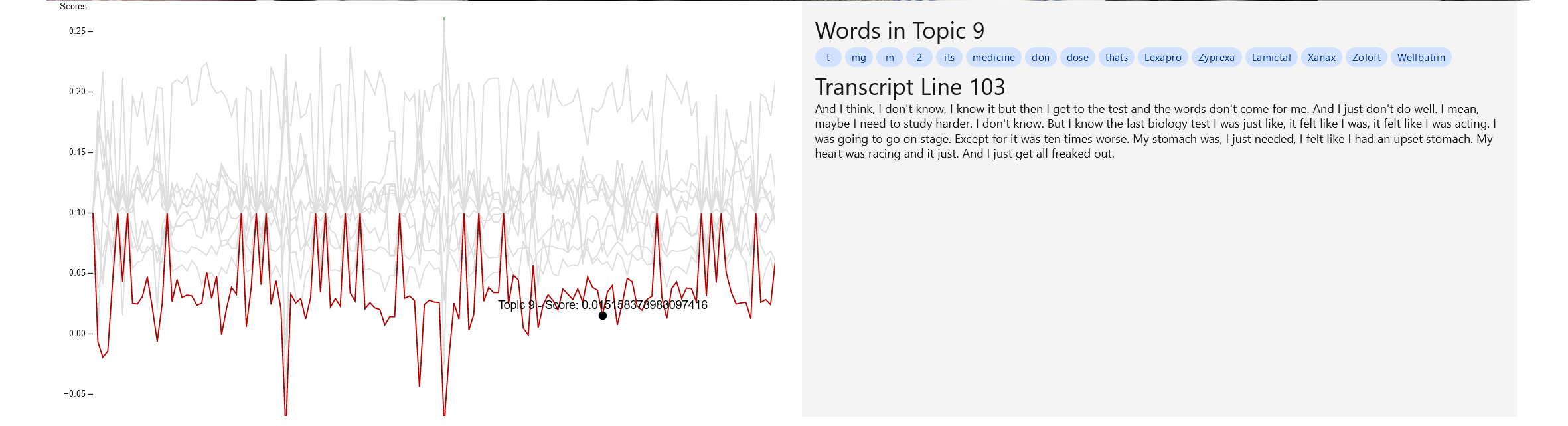}
\caption{Screenshot of the highlighted details in the web dashboard
}\label{fig:screenshot2}
\end{figure}

\section{The TherapyView Demonstration System}

\subsection{The web dashboard}

These metrics and insights are visualized in a web dashboard (Figure \ref{fig:screenshot}). It consists of two parts: a Jupyter notebook that generates and serves the data, and a visual interactive dashboard that displays the data. There are four different visualizations in the dashboard:

\begin{itemize}
\item Images as a visual summary of this therapy session, powered by OpenAI's DALL-E 2 API.
\item Line graph of topical tendency over time (as the dialogue turns), which is highlightable for more details (Figure \ref{fig:screenshot2}).
\item 3D plot showing the relationship of the selected three out of the ten topics over time.
\item Full readout of the transcript (which can be replaced with user-specified inputs).
\end{itemize}

Each AI-generated image represents a single 1,000 character excerpt from the loaded transcript. These images act as a visual timeline, potentially surfacing notable changes in the patient during a session. The vague nature of these images is supplemented by the numerical data provided by the neural topic model. The therapist can explore each of the topics in detail through the charts described above. If the therapist finds a topic score change of interest, they can retrieve the corresponding line in the transcript and analyze the raw text.

This dashboard allows therapists to identify elements of concern by presenting them visually. By quickly identifying these elements, a therapist can provide the appropriate treatment in a timely fashion. These visualizations may also help surface behaviors that may remain un-noticed by the therapist with out the help of the dashboard.  

\subsection{The system architecture}

The system architecture of the dashboard consists of two components: an API and a web application. The API is a single Jupyter notebook written in Python. This notebook contains all the logic for generating the visualizations in the dashboard. The ``Jupyter Kernel Gateway'' package turns each cell into an API endpoint. The web component is a React single page application the queries the API for the data, displays it, and adds interactivity.

This separation of concerns allows scientists to quickly iterate and experiment. Jupyter notebooks are generally understood by AI researchers, and it does not require any special knowledge for them to add new features to the API. 

This is a design optimized for rapid prototyping and experimentation. It is not production ready. Commercialization of this dashboard will require that the Jupyter notebook be replaced with a more robust, efficient, and permanent solution. 

\subsection{The limitations of the AI image generation}

Out of all the visualizations on the dashboard, the generated images are of special interest. Every refresh of the dashboard generates a new set of images, making the results unpredictable. This was included because novel AI approaches like DALL-E, even if they are imprecise, has the potential to provide new perspectives for a therapist to consider.

Integrating DALL-E with real-world therapy does have some challenges:
 (1) the API only allows a maximum of 1,000 characters per image request. This means that DALL-E cannot use any context outside of small chunks, which may limit the kind of insights that it has the potential to visualize;
(2) in the demo, a number of prompts were rejected by DALL-E for ``safety'' reasons. OpenAI prevents certain topics to be visualized for ethical reasons. Psychotherapy sessions can involve many sensitive topics and harmful behaviors. Further development of this approach will require many safeguards to ensure privacy and ensure ethical use.

\section{Conclusions}

In conclusion, this data visualization demonstration system presents a visual journey through the doctor-patient dialogues in therapy sessions via temporal topic modeling and image generation. The results of our study show that the Embedded Topic Model yields high topic coherence and diversity, making it a strong candidate for use in this domain. Our incorporation of temporal modeling and interactive modules on the web dashboard provide additional interpretability, allowing therapists to better understand the progression of psychiatric conditions over time. The use of AI-generated artworks further enhances the interpretability of the results, providing therapists with a visual representation of the core themes of a given therapy session. The results of this study provide valuable insights into the session trajectories of patients and therapists and have the potential to improve the effectiveness of psychotherapy. This is just the first step in a potential turn-level resolution temporal analysis of topic modeling, and we look forward to further exploring this area in future studies.

In future work, we plan to use the learned topic scores to predict psychological or therapeutic states with other digital traces \cite{lin2020predicting,lin2022predicting}. Additionally, we aim to train chatbots as reinforcement learning agents using these states, incorporating biological and cognitive priors \cite{lin2020story,lin2020unified,lin2019split,lin2021models}, and studying their factorial relations with other inference anchors, such as working alliance and personality \cite{lin2022deep,lin2022deep2,lin2022unsupervised}. Our ultimate goal is to construct a complete AI knowledge management system for mental health, utilizing different NLP annotations in real-time (as in review \cite{lin2022knowledge}), and drive AI-augmented therapy sessions \cite{lin2022supervisor,lin2022supervisor2}.


In conclusion, our proposed TherapyView system represents a novel approach to psychotherapy, leveraging the latest advancements in deep learning and data visualization to help therapists provide better care for their patients. The use of NLP and AI-generated arts in our system enables therapists to quickly identify patterns in patient data and tailor their treatment strategies accordingly. With further development and testing, TherapyView has the potential to revolutionize the field of psychiatry and improve patient outcomes.

\bibliography{main}
\bibliographystyle{unsrt}

\end{document}